\documentclass[sigconf]{acmart}
\pdfoutput=1
\usepackage{multirow}
\usepackage{graphicx}
\usepackage{paralist} 
\usepackage{amsfonts}
\usepackage{amsmath}
\usepackage{amsthm}
\usepackage{makecell}
\usepackage{subcaption}
\usepackage{enumitem}
\usepackage[normalem]{ulem} 
\usepackage{url}
\usepackage{lineno}

\AtBeginDocument{%
  \providecommand\BibTeX{{%
    \normalfont B\kern-0.5em{\scshape i\kern-0.25em b}\kern-0.8em\TeX}}}

\copyrightyear{2020}
\acmYear{2020}
\setcopyright{iw3c2w3}

\acmConference[WWW '20]{Proceedings of The Web Conference 2020}{April 20--24, 2020}{Taipei, Taiwan}
\acmBooktitle{Proceedings of The Web Conference 2020 (WWW '20), April 20--24, 2020, Taipei, Taiwan}
\acmPrice{}
\acmDOI{10.1145/3366423.3380223}
\acmISBN{978-1-4503-7023-3/20/04}

\begin{document}

\title{What Changed Your Mind: The Roles of Dynamic Topics and Discourse in Argumentation Process}

 

\author{Jichuan Zeng}
\affiliation{%
  \institution{The Chinese University of Hong Kong}
  \city{Hong Kong}
  \country{China}
}
\email{jczeng@cse.cuhk.edu.hk}

\author{Jing Li}
\affiliation{%
  \institution{The Hong Kong Polytechnic University, Hong Kong, China}
}
\email{jing-amelia.li@polyu.edu.hk}

\author{Yulan He}
\affiliation{%
  \institution{University of Warwick}
  \country{United Kingdom}
}
\email{yulan.he@warwick.ac.uk}

\author{Cuiyun Gao}
\authornote{\authornotemark[1] Cuiyun Gao is the corresponding author.}
\affiliation{%
  \institution{Harbin Institute of Technology (Shenzhen), China}
}
\email{gcyydxf@gmail.com}

\author{Michael R. Lyu}
\affiliation{%
  \institution{The Chinese University of Hong Kong}
  \city{Hong Kong}
  \country{China}
}
\email{lyu@cse.cuhk.edu.hk}

\author{Irwin King}
\affiliation{%
  \institution{The Chinese University of Hong Kong}
  \city{Hong Kong}
  \country{China}
}
\email{king@cse.cuhk.edu.hk}

\renewcommand{\shortauthors}{Zeng, et al.}

\begin{abstract}

In our world with full of uncertainty, debates and argumentation contribute to the progress of science and society.
Despite of the increasing attention to characterize human arguments, most progress made so far focus on the debate outcome, largely ignoring the dynamic patterns in argumentation processes. 
This paper presents a study that automatically analyzes the key factors in argument persuasiveness, beyond simply predicting who will persuade whom.
Specifically, we propose a novel neural model that is able to dynamically track the changes of latent \textit{topics and discourse} in argumentative conversations, 
allowing the investigation of their roles in influencing the outcomes of persuasion.
Extensive experiments have been conducted on argumentative conversations on both social media and supreme court.
The results show that our model 
outperforms state-of-the-art models in identifying persuasive arguments via explicitly exploring dynamic factors of topic and discourse.
We further analyze the effects of topics and discourse 
on persuasiveness, and find that they are both useful --- topics provide concrete evidence while superior discourse styles may bias participants, especially in social media arguments.
In addition, we draw some findings from our empirical results, which will help people better engage in future persuasive conversations.
\end{abstract}




\maketitle

\section{Introduction}

\textit{``The aim of argument, or of discussion, should not be victory, but progress.'' --- Joseph Joubert} 

\textbf{Argumentation process} is a turn-taking dialogue mostly held to increase the acceptability of a controversial standpoint.
In the process, a series of connected propositions (henceforth \textbf{arguments}) are put forward intending to justify or refute a standpoint before a rational judge~\cite{van2013fundamentals}.
It plays an essential role in making decisions, constructing knowledge, and bringing truths and better ideas to life~\cite{DBLP:conf/naacl/JoPJSRN18}.
Consequently, the understanding of argumentation processes will help individuals and human society better engage with conflicting stances and open up their minds to pros and cons~\cite{leary2017cognitive}.
It collides different ideas to form thoughts and knowledge, contributing to advance science and society forward~\cite{willard1996liberalism}. 
However, making sense of argumentative conversations is a daunting task for human readers, mostly due to the varied viewpoints and evidence continuously put forward and the complicated interaction structure therein; not to mention 
huge volume of 
argumentation data appearing 
on online platforms every day.

We hence study how to automatically understand argumentation processes, predicting who will persuade whom and figuring out why it happens. 
To date, much progress made in persuasiveness prediction has focused on individual arguments, the wordings therein~\cite{DBLP:conf/acl/WeiLL16,DBLP:conf/acl/HabernalG16}, and how they locally connect with other arguments~\cite{DBLP:conf/argmining/HideyMHMM17,DBLP:conf/coling/JiWHLZH18}.
On the contrary, we examine the context and the dynamic progress of argumentative conversations, which is beyond the studies of argument-level persuasiveness.
Some 
research work analyze 
argument interactions~\cite{DBLP:journals/tacl/WangBSQ17, DBLP:conf/aaai/HideyM18}  to predict who will win the debate. 
Most of them focus on the outcome of argumentation instead of diving deep into the argumentation process~\cite{DBLP:conf/www/TanNDL16,DBLP:conf/naacl/JoPJSRN18}.
The latter, however, is arguably the essence of argumentation, revealing how participants collaborate to reshape and refine ideas.

In light of these missing points, we track the argumentation process and explicitly explore the dynamic patterns of what a discussion is centered around (henceforth \textbf{topics}) and how the participants voice their opinion in arguments (henceforth \textbf{discourse}), as well as how they affect the persuasion results.
To illustrate the interplay of topics and discourse in argument persuasiveness, Figure \ref{fig:intro-example} shows a Reddit conversation snippet from ChangeMyView subreddit.\footnote{\url{https://www.reddit.com/r/changemyview/}}
On ChangeMyView, an opinion holder first raises a viewpoint (henceforth $OP$ short for original post), followed by challengers' arguments attempting to change the opinion holder's mind.
This example dialogue is formed with challengers' arguments against ``\textit{learning a second language isn't worth it for most people anymore}'', which was the opinion holder's point of view.

\begin{figure}[t]
    \centering
    \small
    \begin{tabular}{|m{7cm}|}
    \hline
    $OP$: Translation software already exists and is pretty good. ... Most people putting in years of effort to learn a foreign language skill they might only use a couple times in their whole life, and will likely forget.
    \\
    $A_1$ $[$\textit{Evidence}$]$: ... There is research that indicates ``that \textcolor{red}{\emph{those who spoke two or more languages had significantly better cognitive abilities compared to what would have been expected from their baseline test}}.'' $\langle$url$\rangle$.
    ... Another study found that `` \textcolor{red}{\emph{the language-learning participants ended up with increased density in their grey matter and that their white matter tissue had been strengthened. }}'' $\langle$url$\rangle$\\
    $A_2$ $[$\textit{Metaphor}$]$: The common comparison is made to learning music, as /u/awesomeosprey has pointed out. I did some research into the matter. It seems that \textcolor{red}{\emph{learning a musical instrument does have long-lasting benefits} ($\langle$url$\rangle$)  \emph{that relate to ``higher-order aspects of cognition.}}'' \\
    ...\\
    $A_4$ $[$\textit{Reference}$]$ ... But a quick search and I have other sources: $\langle$digit$\rangle$ $\langle$url$\rangle$, $\langle$digit$\rangle$ $\langle$url$\rangle$, $\langle$digit$\rangle$ $\langle$url$\rangle$. 
    The most interesting study is this one ($\langle$url$\rangle$), but I can't find a complete version of it, sorry. /n/nNote: Study $\langle$digit$\rangle$ has an exceptionally small sample size. It's still interesting reading.\\
    \hline
    \end{tabular}
    \caption{A ChangeMyView conversation snippet of challengers' arguments against $OP$ raised by the opinion holder concerning ``\textit{learning a second language isn't worth it anymore for most people}''. The red and italic words indicate the key points resulting in the challengers' victory. The words in $[]$ are our interpretations of the arguments' discourse styles.}
    \label{fig:intro-example}
\end{figure}

It is seen that the challengers successfully persuaded the opinion holder to change their view in the aforementioned example. The probable reasons are two fold. First, there are strong evidences (reflected by topic words) put forward, such as the research findings on cognitive abilities. Second, they deploy skillful debating styles (captured by discourse words), such as the metaphors with learning music (in $A_2$) and the reference to external information (in $A_4$).

Motivated with these observations, we propose a novel neural framework that explicitly models how the change of discussion topic and discourse styles affect persuasion effectiveness.
Our model first explores latent topics and discourse in arguments with word clusters. 
Furthermore, it tracks topic change and discourse flow in the argumentation process and automatically interprets the key factors indicating the success or failure of the persuasion.
Coupling the advantages of neural topic models~\cite{DBLP:conf/icml/MiaoGB17,DBLP:conf/emnlp/ZengLSGLK18,DBLP:journals/corr/abs-1903-07319} and dynamic memory networks~\cite{DBLP:conf/icml/KumarIOIBGZPS16,DBLP:conf/icml/XiongMS16,DBLP:conf/www/ZhangSKY17}, we are able to explore dynamic topic and discourse representations indicative of persuasiveness in an end-to-end manner with the persuasion outcome prediction. 
To the best of our knowledge, we are the first to \emph{explicitly model topics and discourse in argumentation processes, and investigate how their dynamic patterns contribute to the argument persuasiveness.}

We carry out extensive experiments on argumentative conversations gathered from both social media and U.S. supreme court. 
The results show that our model can significantly outperform  state-of-the-art methods on both datasets, which shows its effectiveness in identifying persuasive arguments. 
For example, we achieve $70.2$\% accuracy when predicting winners in supreme court debates, compared with $63.1$\% obtained by logistic regression without explicitly exploiting dynamic topics and discourse features in argumentation processes.
Based on the produced topics and discourse, we further analyze how they affect persuasiveness. 
It is indicated that topics (such as evidence and viewpoints) statistically contribute more on persuasion success while skillful discourse style may sometimes lead to victory.
In addition, we summarize the key findings 
from our empirical results, which will help individuals 
better engage in future persuasions.

To sum up, our contributions are three folds:
\begin{itemize}
    \item We are the first to study the argumentation process via dynamic analysis of latent topics and discourse, which reveals the key factors in argument persuasiveness. 
    \item We propose a novel neural model to predict argumentation outcome via tracking dynamic topic and discourse patterns in the dialogue process.
   
    \item We provide an extensive empirical study on two real-world datasets that demonstrates the effectiveness of our model and sheds light on a better understanding and development of persuasive augmentations.
\end{itemize}

\section{Related Work}\label{sec:related}


\subsection{Argument Persuasiveness} 
As a fast growing sub-field of computational argumentation mining~\cite{DBLP:conf/emnlp/StabG14,DBLP:conf/acl/WachsmuthNHHHGS17}, previous work in this area mostly work on the identification of convincing arguments~\cite{DBLP:conf/acl/WeiLL16,DBLP:conf/acl/HabernalG16} and viewpoints~\cite{DBLP:conf/aaai/HideyM18,DBLP:conf/naacl/JoPJSRN18} from varying argumentation genres, such as social media discussions~\cite{DBLP:conf/www/TanNDL16}, political debates~\cite{DBLP:conf/naacl/BasaveH16}, and student essays~\cite{DBLP:conf/acl/NgCGK18}.
In this line, many existing studies focus on crafting hand-made features~\cite{DBLP:conf/acl/WeiLL16,DBLP:conf/www/TanNDL16}, such as wordings and topic strengths~\cite{DBLP:conf/naacl/Zhang0RD16,DBLP:journals/tacl/WangBSQ17}, echoed words~\cite{atkinson+srinivasan+tan:19}, semantic and syntactic rules~\cite{DBLP:conf/argmining/HideyMHMM17,DBLP:conf/ijcai/PersingN17}, participants' personality~\cite{DBLP:conf/eacl/WalkerALW17}, argument interactions and structure~\cite{DBLP:conf/acl/NiculaePC17}, and so forth. 
These methods, however, 
require labor-intensive feature engineering process, and hence have limited generalization abilities to new domains. 

Recently, built upon the success of neural models in natural language processing (NLP), neural argumentation mining methods have been proposed
to enable end-to-end learning of automatic features and argument persuasiveness. 
For example, \citet{DBLP:journals/corr/PotashRR16b} tailor a pointer network architecture to learn argument representations. 
\citet{DBLP:conf/argmining/LinHHC19} focus on incorporating external lexicons into an attentive neural network for argumentative component identification. 
These studies, however, ignore the dynamic nature of argumentation process, where the persuasion features may change in a heated back-and-forth debate.
Some other methods consider the modeling of the argument interactions in persuasiveness prediction. 
\citet{DBLP:conf/coling/JiWHLZH18} explore the argument-level interactions between ChangeMyView original post ($OP$) and its following comments with a co-attention network. 
\citet{DBLP:conf/naacl/JoPJSRN18} investigate the interplay between $OP$ and its challenger's argument, explicitly identifying the amenable parts of $OP$ that is likely to be affected with good arguments. 
Compared with these work focusing on interaction between $OP$ and comments, we dynamically track the entire argumentation flow and capture how topics and discourse therein change and affect persuasion outcomes.
\citet{DBLP:conf/aaai/HideyM18} employ sequence modeling to learn implicit persuasiveness signals from chronologically ordered arguments.
Different from them, we explicitly capture the dynamic topics and discourse behaviors as discussion process is moved forward, where their roles in shaping the persuasive arguments can be examined. 

\subsection{Conversation Process Understanding}
Our work is also closely related with conversation process understanding. 
In this line, previous studies have shown the benefits of 
discovering the latent discourse structure.
It shapes how utterances interact with each other and form the discussion flow with the use of dialogue acts (e.g., making a statement, asking a question, and giving an example).
Most of them extend Hidden Markov Model (HMM) to produce distributional clusters of words to reflect latent discourse ~\cite{chotimongkol2008learning,DBLP:conf/naacl/RitterCD10}.
In discourse learning, features are exploited via modeling of conversation tree structure~\cite{DBLP:journals/coling/LiSWW18}, relative position of sentences~\cite{DBLP:conf/ijcai/JotyCL11}, topic content~\cite{DBLP:conf/acl/ZhaiW14,DBLP:conf/acl/QinWK17}, and so forth.


In addition, the recent progress in recurrent variational neural networks (non-linear HMM counterpart) enables to capture latent discourse structure in dialogues.
For example, latent variable RNN (LVRNN) and variational RNN (VRNN) have been adopted to model the latent conversation states in each turn~\cite{DBLP:journals/corr/JiHE16,DBLP:conf/naacl/ShiZY19}. 
\citet{DBLP:journals/corr/abs-1903-07319} jointly explore the topic content and discourse behavior to better understand conversations  
by using the word clusters to represent topics and discourse in microblog conversations. 
However, none of them captures how topics and discourse change in a conversation process and how these dynamic patterns affect argumentation persuasiveness, which is the gap our work fills in.
\section{Study Design}\label{sec:data}
In this section, we first introduce how we formulate our problem, followed by a detailed discussion on the experimental datasets. 

\subsection{Problem Formulation}

In this paper, we define \textbf{argumentation process} $C$ as a dynamic conversation process held by participants.
It is formulated as a sequence of turns, denoted as $C=\{x_t\}_{t=1}^{T}$, where a turn $x_t$ refers to an argument and $T$ the number of turns in the process. 
As discussed above, our work studies argument persuasiveness in the context of its discussion process, which however relies on subjective judgement.
After all, human performance on ``yes-or-no'' persuasiveness judgement is still close to random guess~\cite{DBLP:conf/www/TanNDL16}.
In our study, we view argument persuasiveness from a perspective of comparison
(instead of answering ``yes or no''), and formulate its prediction as a pair-wise ranking problem under a debate $\mathcal{D}$.
Concretely, we construct the pairwise comparison settings to take a pair of argumentation process $\langle C_i, C_j \rangle$ as input,
where $C_i, C_j \in \mathcal{D}$; Scores $y_i$ and $y_j$ are assigned to measure their persuasiveness respectively.
Here $y_i > y_j$ means that $C_i$ has a better chance to win the debate compared with $C_j$, while $y_i < y_j$ otherwise.
The goal of our paper is to predict which argumentation process from the input pair is relatively more persuasive and analyze the key factors therein to reveal insights for argumentation study.
\begin{figure}[t]
	\centering
	\begin{subfigure}[h]{0.23\textwidth}
	\includegraphics[width=0.98 \textwidth]{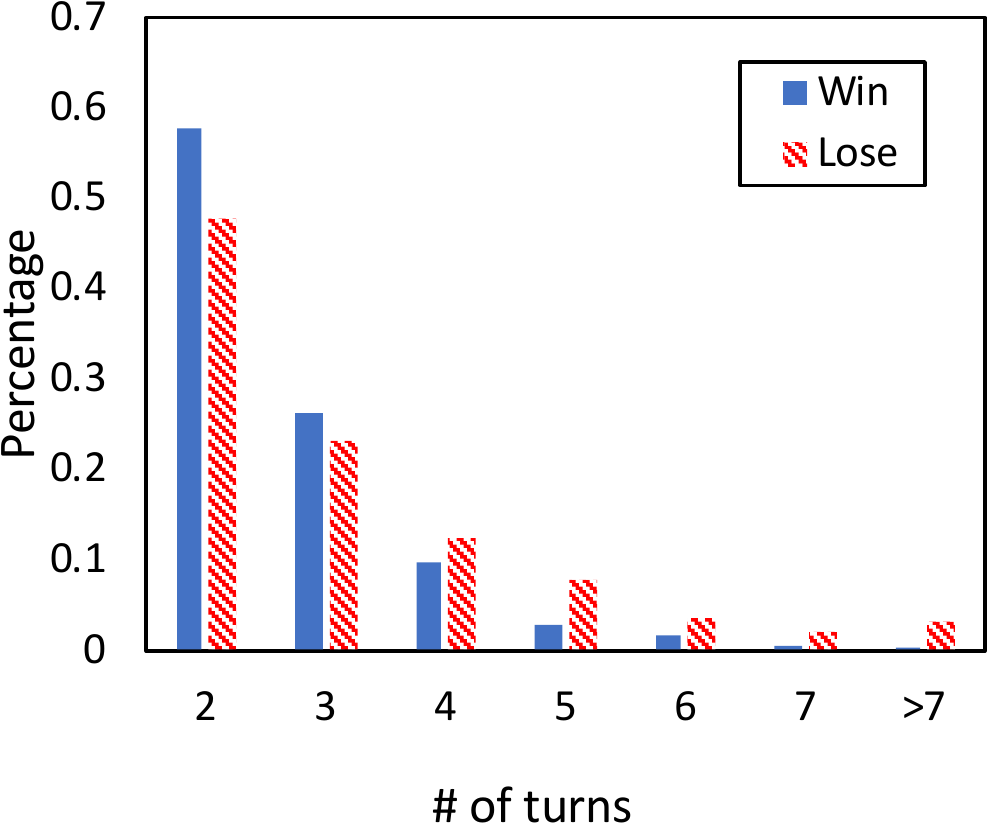}
    \caption{CMV}
    \end{subfigure}
    \hfill
    \begin{subfigure}[h]{0.23\textwidth}
    \includegraphics[width=0.98 \textwidth]{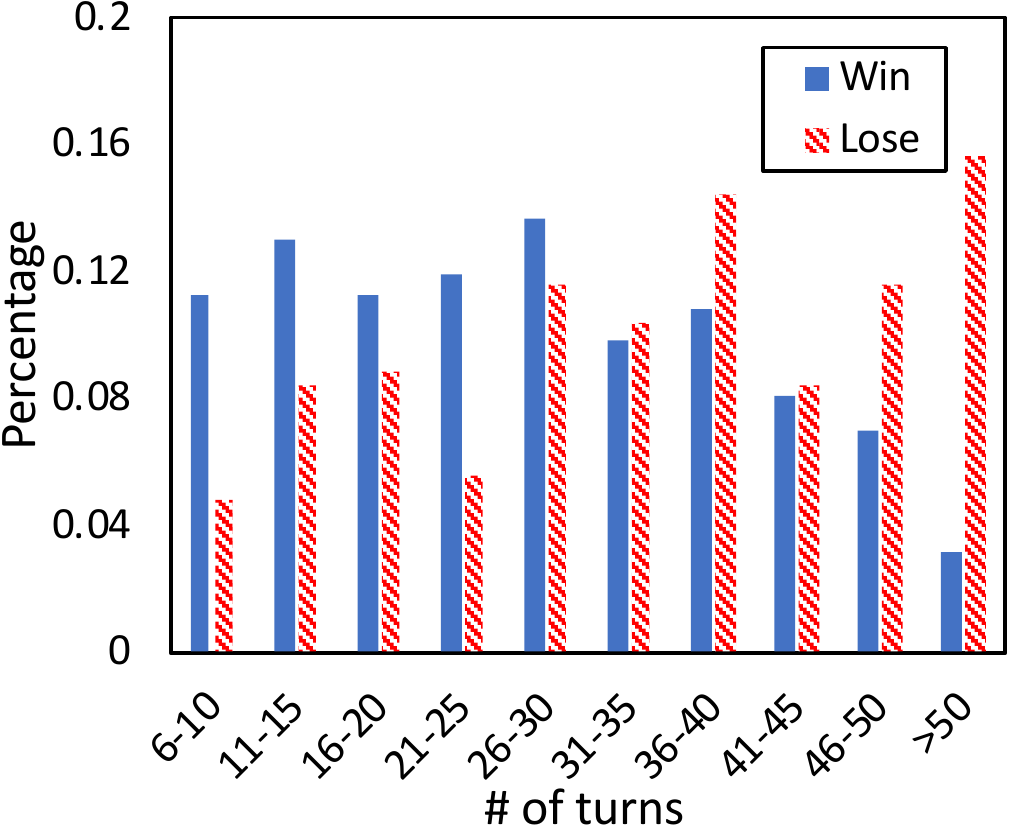}
    \caption{Court}
    \end{subfigure}
    \vspace{-0.25cm}
    \caption{Distributions over the wining and losing augmentation processes concerning their number (range) of turns. (a) is for ChangeMyView (CMV) dataset and (b) supreme court (Court) dataset.
    }
    \vspace{-0.25cm}
\label{fig:turn_lens}
\end{figure}

Our problem setting can fit diverse scenarios to learn what a good persuasion should be. 
For example, it works for the classic Oxford-style debate involving two sides, where one argues ``for'' a statement and the other ``against''.
The arguments from both sides can be defined as $C^f =\{x_t^f\}$ (``for'' side) and $C^a=\{x_t^a\}$ (``against'' side), which corresponds to our input pair in problem setting.

\begin{table}[t]\small
    \centering
    \caption{Statistics of the ChangeMyView (CMV) and the supreme court (Court) datasets. Here a moot refers to an original post in CMV and a case in Court.}\label{tab:statistics}
    \vspace{-0.25cm}
     \scalebox{0.95}{
    \begin{tabular}{|l|rrrrrr|}
         \hline
         \multirow{2}{*}{\textbf{Datasets}}& \# of &  \# of & \# of & avg. words&\multirow{2}{*}{|vocab|} & \# of \\
         &moots&convs&turns&per turn& & pairs\\
         \hline
         CMV & 2,396 & 30,341 & 109,644 & 96.2 & 13,541 & 12,879 \\
         Court & 204 & 655 & 17,599 & 46.1 & 6,260 & 3,656 \\
         \hline
    \end{tabular}
    }
    \vspace{-0.25cm}
\end{table}

\subsection{Data Description}
We conduct our study in two scenarios --- social media arguments, which tend to use colloquial and informal languages, 
and supreme court debates,\footnote{\url{https://www.supremecourt.gov/oral_arguments/oral_arguments.aspx}} exhibiting a more formal language style.
The social media arguments are gathered from the ChangeMyView subreddit, where challengers engage in the discussion with  attempts to change the opinion holder's view (pointed out in the original post $OP$)~\cite{DBLP:conf/www/TanNDL16}.
As a multi-party conversation, a debate there is in tree structure formed with in-reply-to relations (a post can have multiple replies), and a path therein is defined as an argumentation process. 
We aim to predict which path has a better chance to be awarded a $\Delta$
by the opinion holder to indicate successful persuasion.
For the supreme court debates, we aim to predict whether the petitioner or respondent will win the case, given their corresponding conversational exchanges with the justices.

The ChangMyView social media dataset (henceforth \textbf{CMV}) is built with a corpus released by \citet{DBLP:conf/www/TanNDL16} with argumentative conversations held from Jan 2013 to May 2015. 
As stated above, each discussion in CMV can be organized in a tree structure with in-reply-to relations (henceforth a debate tree), with its root representing the $OP$ (the opinion holder's viewpoint). 
To construct our input data, following~\citet{DBLP:conf/www/TanNDL16}, we first filter out the trivial cases by removing the discussions with less than $10$ challengers, or those do not contain a $\Delta$. 
Then, we flatten the debate tree into conversation paths and remove replies with $50$ words or less.
Also removed are conversation paths involving less than two turns\footnote{In our paper, unless 
otherwise specified, a \textbf{conversation} is used as the short form of a conversation path.}.
Next, all challengers' replies remained in a conversation path is considered as the turns in argumentation process.
For each debate tree, we form a positive candidate set with all the argumentation processes (paths) leading to a $\Delta$, and include those without a $\Delta$ into the negative candidate set.
To formulate our pairwise inputs, we perform the Cartesian product\footnote{\url{https://en.wikipedia.org/wiki/Cartesian_product}} on the positive and negative candidate sets, which returns all the possible combinations of successful-unsuccessful argumentation process pairs in the debate.

For the supreme court debate dataset (abbreviated as \textbf{Court}), it is gathered by \citet{DBLP:conf/www/Danescu-Niculescu-MizilLPK12} from the U.S. supreme court dialogues\footnote{\url{http://www.supremecourt.gov/oral_arguments/}}. In this corpus, the petitioner and respondent make conversational exchanges to justices to defend for themselves in turn. 
Here the petitioner's utterances are taken to form its augmentation process, and so does the respondent's.
For each case, we build the positive candidate set with argumentation processes from the wining side, and negative from its opponent. 
The pairwise inputs are formed following the similar procedure used for CMV dataset.

\begin{figure}[t]
	\centering
	\includegraphics[width=0.32 \textwidth]{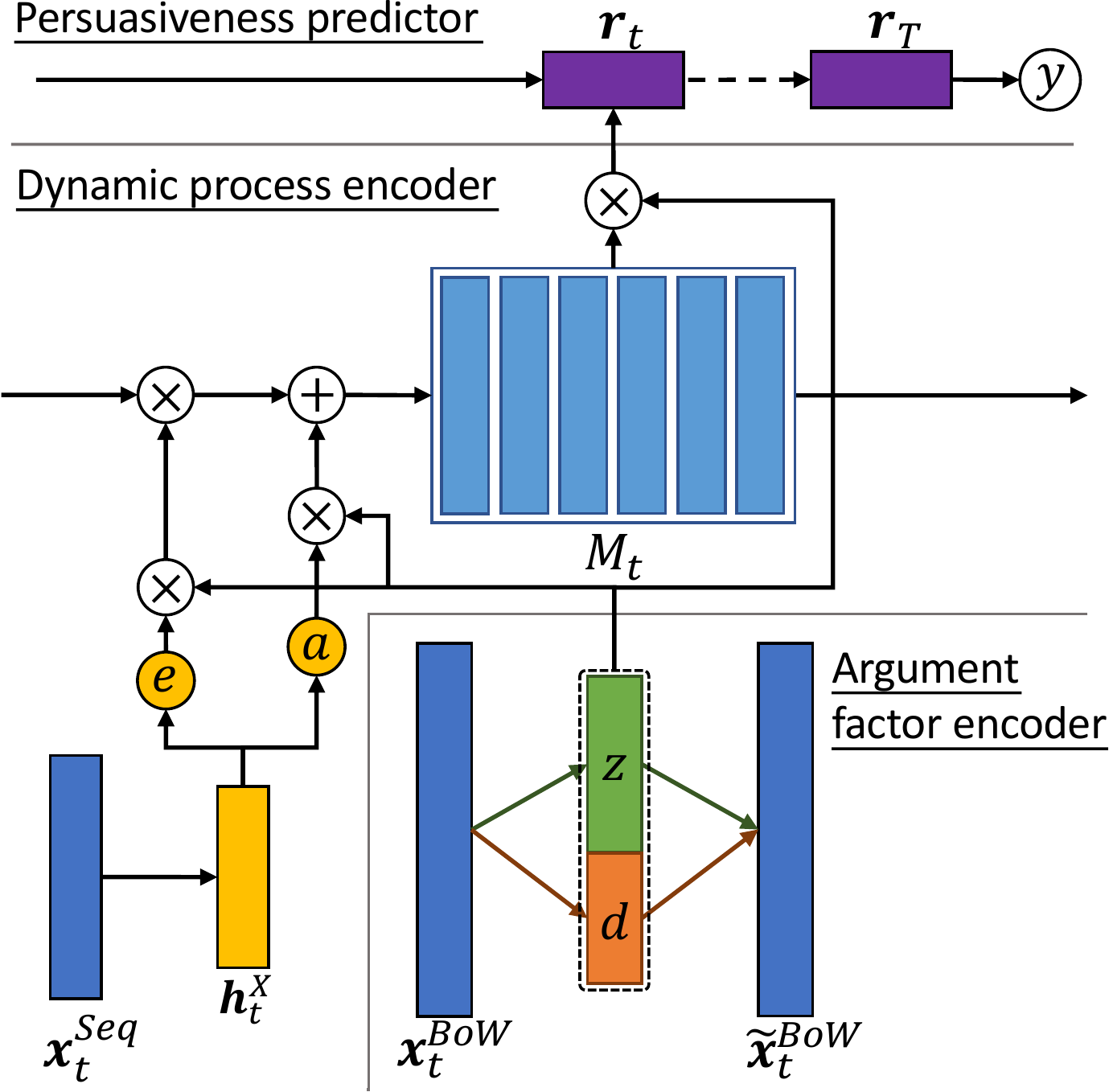}
\vskip -0.5em
	\caption{
    The architecture of our dynamic topic-discourse memory networks (DTDMN) for persuasiveness prediction. 
    }
	\label{fig:framework}
    \vskip -1em
\end{figure}

In addition, we employ two strategies to further improve the quality of our data. 
First, to ensure the argumentation processes in an input pair concern relevant topics, their Jaccard similarity are measured over bag-of-words form.
After that, following the practice in~\citet{DBLP:conf/www/TanNDL16}, we remove pairs with $<0.5$ Jaccard similarity where the  conversation pairs may not be on the same page.
Second, as pointed out in previous studies~\cite{DBLP:conf/www/TanNDL16}, the number of argument turns can largely affect the debate outcome.
Here we show the distribution of turn number over winning and losing argumentation processes in Figure~\ref{fig:turn_lens} and observe that the wining ones tend to be shorter (with smaller turn number).
It might result in trivial features of turn number to be learned for persuasiveness prediction.
To mitigate the effects of turn number and better study the roles of topics and discourse, we make sure that the pairwise processes fed to the model are equally long (have the same number of argument turns).
To this end, we remove pairs with shorter negative process and for the rest, we truncate the longer parts of negative processes. 
The statistics of our two datasets are shown in Table~\ref{tab:statistics}. As can be seen, there are more conversations in CMV than Court. However, the Court debates involve more turns ($26.9$ vs. $3.6$ turns on average per conversation). 
It might be because court debates are more serious and usually result in a back-and-forth fashion while social media discussions are mostly casual and may end soon.

It is worth noting that we do not feed the words from either opinion holders or justices to avoid the possible bias incurred in persuasiveness prediction. 
In doing so, we can 
focus on linguistic features in participants' arguments that lead to good persuasion. Further, 
it enables our setting to be easily adapted to scenarios without the third-party engagement (e.g., opinion holders and justices).
In addition, for CMV dataset, we consider the engagements of all challengers regardless of their $\Delta$ records, which is different from the setting in \citet{DBLP:conf/www/TanNDL16}, which only examine the $\Delta$ winners.
It is because everyone's efforts may contribute to the final success (or failure) of an argumentation process.
Therefore, all the challengers' argument are taken into account in our persuasiveness analysis.
For the same reason, we have more training data instances than those in~\citet{DBLP:conf/www/TanNDL16} ($12,879$ vs. $4,263$).

\section{DTDMN: Dynamic Topic-Discourse Memory Networks for Argument Persuasiveness}\label{sec:approach}
This section presents our model that predicts persuasiveness, and dynamically discovers the key topic and discourse factors therein to explain the reasons behind. 
Our model, named as dynamic topic-discourse memory networks (DTDMN), consists of three modules --- one to learn latent topic and discourse factors from each argument (henceforth \textbf{argument factor encoder}), one to explore the change of topic and discourse factors in argumentation flows (henceforth \textbf{dynamic process encoder}), and the last one 
to identify the more persuasive conversation from the input pair (henceforth \textbf{persuasiveness predictor}).
The model architecture is shown in Figure \ref{fig:framework} with an overview presented in Section \ref{ssec:model}. Then 
in Section \ref{ssec:argument-factor-encoder}, \ref{ssec:dynamic-process-encoder}, and \ref{ssec:persuasiveness_predictor}, we describe our three modules in turn, followed by our learning objective discussed in Section \ref{ssec:learning-objective}.

\subsection{Model Overview}\label{ssec:model}
As described in Section \ref{sec:data},
our model takes pairwise conversations as input. 
In training, we feed $\langle\boldsymbol{C}^{+}; \boldsymbol{C}^{-} \rangle$ into our model, where $\boldsymbol{C}^{+}$ is a positive instance referring to a persuasive conversation. Likewise, $\boldsymbol{C}^{-}$, the negative instance, denotes a failed persuasion. During the testing, 
given two conversations, our model will recognize the one which is more persuasive. 
Each conversation $\boldsymbol{C}$ is formed with a sequence of argumentative turns (henceforth \textbf{arguments}): $\boldsymbol{C}=\langle\boldsymbol{x}_1, \dots, \boldsymbol{x}_T\rangle$, where $T$ denotes the number of arguments in $\boldsymbol{C}$. 

For the $t$-th argument $\boldsymbol{x}_t$, we capture argument-level representations, $\boldsymbol{z}_t\in\mathbb{R}^K$ for topic factor and $\boldsymbol{d}_t\in \mathbb{R}^D$ for discourse factor, from the input of bag-of-words vector $\boldsymbol{x}_t^{BoW}\in \mathbb{R}^V$, where $K$ is the number of topics, $D$ discourse, and $V$ the vocabulary size. 
Then, $\boldsymbol{z}_t$ and $\boldsymbol{d}_t$ are fed into the dynamic memory, together with the word index sequence $\boldsymbol{x}_t^{Seq}\in \mathbb{R}^L$, to update the memory state, where $L$ is the sequence length. 
The output of the dynamic memory networks is used to predict the persuasiveness score $y$ for each conversation, where higher scores  indicate better persuasiveness. Our training target is to have $y^+>y^-$ for $\boldsymbol{C}^{+}$ and $\boldsymbol{C}^{-}$.

\subsection{Argument Factor Encoder}\label{ssec:argument-factor-encoder}
This section presents how we capture topic and discourse factors at the 
argument level.
The subscript $t$ is omitted for simplicity.
As mentioned in Section \ref{ssec:model}, we employ latent variables $\boldsymbol{z}$ for argument topic factor representation, and $\boldsymbol{d}$ for discourse. 
The modeling process is inspired by \citet{DBLP:journals/tacl/abs-1903-07319} and based on variational auto-encoder (VAE)~\cite{DBLP:journals/corr/KingmaW13} to reconstruct a given argument in the BoW form, $\boldsymbol{x}^{BoW}$, conditioned on 
$\boldsymbol{z}$ and $\boldsymbol{d}$. 
Here $\boldsymbol{z}$ is the topic mixture and $\boldsymbol{d}$ is a one-hot vector denoting the discourse style.\footnote{We follow the setting of~\citet{DBLP:journals/tacl/abs-1903-07319}, and apply Gumbel-Softmax relaxation for $\boldsymbol{d}$.}
Specifically, the generation process for each word $w_n\in \boldsymbol{x}^{BoW}$ is defined as:
\begin{equation}
\begin{aligned}
&\boldsymbol{\epsilon}\sim\mathcal{N}(\boldsymbol{\mu}, \boldsymbol{\sigma}^2),\ \ \boldsymbol{z}= \operatorname{softmax}(f_z(\boldsymbol{\epsilon})),\ \ \boldsymbol{d} \sim Multi(\boldsymbol{\pi}),\\
&\beta_n = \operatorname{softmax}(f_{\phi^T} (\boldsymbol{z})+f_{\phi^D} (\boldsymbol{d})), \ \ w_{n}\sim Multi(\beta_{n}),
\end{aligned}
\end{equation}
where $f_*(\cdot)$ is a neural perceptron that linearly transforms inputs.
For both latent topic and discourse factors, we employ word distributions to represent them. 
Here we consider the weight matrix of $f_{\phi^T}(\cdot)$ (after the softmax normalization) as topic-word distributions, $\phi^T$. 
Likewise, $f_{\phi^D}(\cdot)$'s weight matrix is used to compute the discourse-word distributions, $\phi^D$.

For the other parameters $\boldsymbol{\mu}$, $\boldsymbol{\sigma}$, and $\boldsymbol{\pi}$, they can be learned from the input $\boldsymbol{x}^{BoW}$ following the formula below: 
\begin{equation}
\begin{aligned}
\boldsymbol{\mu} = f_{\mu}(\operatorname{tanh}(&f_e(\boldsymbol{x}^{BoW})))
,\, \log\boldsymbol{\sigma} = f_{\sigma}(\operatorname{tanh}(f_e({\boldsymbol{x}^{BoW}}))),
\\
&\boldsymbol{\pi} = \operatorname{softmax}(f_{\pi}(\boldsymbol{x}^{BoW})).
\end{aligned}
\end{equation}

\subsection{Dynamic Process Encoder}\label{ssec:dynamic-process-encoder}

Based on the topic and discourse factors learned at the argument level, here we discuss how to capture their dynamic patterns in the persuasion process.
Our dynamic process encoder is inspired by dynamic memory network (DMN)~\cite{DBLP:conf/icml/KumarIOIBGZPS16,DBLP:conf/icml/XiongMS16,DBLP:conf/www/ZhangSKY17} and topic memory mechanism~\cite{DBLP:conf/emnlp/ZengLSGLK18}, where we capture the indicative dynamic topic and discourse factors to interpret why a conversation can result in successful persuasion.

To be more specific, memory weight $\boldsymbol{w}_t\in \mathbb{R}^{(K+D)}$ is defined as the concatenation of latent aspects $\boldsymbol{z}_t$ and $\boldsymbol{d}_t$:
\begin{equation}
\begin{aligned}
&\boldsymbol{w_t} = [\boldsymbol{z}_t;\boldsymbol{d}_t],
\end{aligned}\label{eq:weight}
\end{equation}
where $[\cdot;\cdot]$ represents 
the concatenation. Once we have the memory weight, DTDMN will retrieve and update the memory according to the memory weight and input argument. 

We employ a bidirectional attentive GRU~\cite{DBLP:journals/corr/BahdanauCB14,DBLP:conf/icml/XuBKCCSZB15} to encode the word index sequence vector input $\boldsymbol{x}_t^{Seq}$ into hidden states $\boldsymbol{h}_t^X \in\mathbb{R}^H$:

\begin{equation}
\begin{aligned}
    \overrightarrow{\boldsymbol{h}}_{t,j}^{X} = \operatorname{\overrightarrow{GRU}}(x_{t,j};& \overrightarrow{\boldsymbol{h}}_{t,j-1}^{X}),\ \ \ 
    \overleftarrow{\boldsymbol{h}}_{t,j}^{X} = \operatorname{\overleftarrow{GRU}}(x_{t,j}; \overleftarrow{\boldsymbol{h}}_{t,j+1}^{X}),
    \\
    \boldsymbol{h}_t^X &= \operatorname{attn}(\{[\overrightarrow{\boldsymbol{h}}_{t,j}^{X};\overleftarrow{\boldsymbol{h}}_{t,j}^{X}]\}_{j=1}^{L}),
\end{aligned}
\end{equation}
where $j\in[1,L]$, $x_{t,j}$ is the $j$-th token in $\boldsymbol{x}_t^{Seq}$. $\operatorname{attn}(\cdot)$ is the attention operator~\cite{DBLP:journals/corr/BahdanauCB14,DBLP:conf/emnlp/LuongPM15} to aggregate the representations of tokens to form a vector representation for $\boldsymbol{x}_t^{Seq}$.

Similar to~\citet{DBLP:conf/www/ZhangSKY17}, we employ a forget gate to erase the retrieved memory. The erase vector is denoted as $\boldsymbol{e}_t\in \mathbb{R}^E$, where $E$ is the dimension of memory embeddings. Afterwards, an augment gate is used to strengthen the retrieved memory. The augment vector is denoted as $\boldsymbol{a}_t\in \mathbb{R}^E$. The overall update formulae for episodic memory are: 
\begin{equation}
\begin{aligned}
&\boldsymbol{M}_{t,i} = \boldsymbol{M}_{t-1,i}[{\bf 1}-w_{t,i} \boldsymbol{e}_t] + w_{t,i} \boldsymbol{a}_t,\\
\boldsymbol{e}_t = \operatorname{sigmoid}&(\boldsymbol{W}^{(e)}\boldsymbol{h}^X_t+\boldsymbol{b}^{(e)}),\ \ 
\boldsymbol{a}_t = \operatorname{tanh}(\boldsymbol{W}^{(a)}\boldsymbol{h}^X_t+\boldsymbol{b}^{(a)}),
\end{aligned}
\end{equation}
where $\boldsymbol{M}_{t,i}\in \mathbb{R}^E$ is the $i$-th row of the memory matrix $\boldsymbol{M}_t$, $\bf 1$ is a row-vector of all $1$s.
$\boldsymbol{W}^{(e)}, \boldsymbol{W}^{(a)}\in \mathbb{R}^{E\times H}$ and $\boldsymbol{b}^{(e)}, \boldsymbol{b}^{(a)}\in \mathbb{R}^{E}$ are the weight matrices and bias vectors for computing $\boldsymbol{e}_t$ and $\boldsymbol{a}_t$, respectively.
%
The read content $\boldsymbol{r}_t\in \mathbb{R}^E$ of the episodic memory $\boldsymbol{M}_t$ is the weighted sum of the memory matrix:
\begin{equation}
\begin{aligned}
&\boldsymbol{r}_t = \sum_{i=1}^{K+D} w_{t,i}\boldsymbol{M}_{t,i}.
\end{aligned}
\end{equation}

\subsection{Persuasiveness Predictor}\label{ssec:persuasiveness_predictor}
For each conversation, DTDMN dynamically summarizes the read contents of the previous arguments in a conversation $\{\boldsymbol{r}_t\}_{t=1}^{T'}$ via an attentive GRU at the argument level:
\begin{equation}
\begin{aligned}
    \boldsymbol{h}_t^R = &\operatorname{GRU}(\boldsymbol{r}_t;\boldsymbol{h}_{t-1}^R),
    \\
    \boldsymbol{h}^R = &\operatorname{attn}(\{\boldsymbol{h}_t^R\}_{t=1}^{T'}).
\end{aligned}
\end{equation}
Then we map $\boldsymbol{h}^R$ to a score value $y$:
\begin{equation}
\begin{aligned}\label{eq:pred_score}
y = \boldsymbol{W}^{(r)}\boldsymbol{h}^R + b^{(r)},
\end{aligned}
\end{equation}
where $\boldsymbol{W}^{(r)}\in\mathbb{R}^{1\times E}$ and $b^{(r)}\in \mathbb{R}^1$ are weight and bias for computing $y$.

\subsection{Learning Objective}\label{ssec:learning-objective}

\vspace{0.5em}
\noindent\textbf{Argument Factor Learning.} To model topic and discourse factors, in learning, we maximize the variational lower bound $\mathcal{L}_{z}$ for $\bf z$ and $\mathcal{L}_{d}$ for $\bf d$. 
The corresponding functions are defined as:

\begin{equation}
\begin{aligned}\label{eq:topic-discourse}
&\mathcal{L}_{z} = \mathbb{E}_{q({\boldsymbol{z}\,|\,\boldsymbol{x}})}[p(\boldsymbol{x}\,|\,\boldsymbol{z})] - D_{KL}(q(\boldsymbol{z}\,|\,\boldsymbol{x})\,||\,p(\boldsymbol{z})),
\\
&\mathcal{L}_{d} = \mathbb{E}_{q(\boldsymbol{d}\,|\,\boldsymbol{x})}[p(\boldsymbol{x}\,|\,\boldsymbol{d})] - D_{KL}(q(\boldsymbol{d}\,|\,\boldsymbol{x})\,||\,p(\boldsymbol{d})),
\end{aligned}
\end{equation}

\noindent where $p(\boldsymbol{z})$ is the standard normal prior $\mathcal{N}({\bf 0}, {\bf I})$ and $p(\boldsymbol{d})$ the uniform distribution $Unif(0,1)$. 
$q(\boldsymbol{z}\,|\,\boldsymbol{x})$ and $q(\boldsymbol{d}\,|\,\boldsymbol{x})$ are  posterior probabilities to approximate how $\boldsymbol{z}$ and $\boldsymbol{d}$ are generated from the arguments. 
$p(\boldsymbol{x}\,|\,\boldsymbol{z})$ and $p(\boldsymbol{x}\,|\,\boldsymbol{d})$ represent the corpus likelihoods conditioned on these topic and discourse factors.

The overall argument factor learning objective is to maxmize:

\begin{equation}\label{eq:final-obj}
\begin{aligned}
\mathcal{L}_{Factor} = \mathcal{L}_{z} + \mathcal{L}_{d} + \mathcal{L}_{x} - \lambda \mathcal{L}_{MI},
\end{aligned}
\end{equation}

\noindent where $\mathcal{L}_{x}$ is for reconstructing the argument $\boldsymbol{x}$ from $\boldsymbol{z}$ and $\boldsymbol{d}$, $\mathcal{L}_{MI}$ is the mutual information (MI) penalty (for separating topic and discourse words). The hyperparameter $\lambda$ is the trade-off parameter for balancing between the $\mathcal{L}_{MI}$ and the other learning objectives. We leave out the details and refer the readers to~\citet{DBLP:journals/tacl/abs-1903-07319}.

\vspace{0.5em}
\noindent\textbf{{Persuasiveness Prediction Learning.}}
In our setting, we aim 
to identify which conversation is more persuasive given an input of two conversations. Therefore, our goal is to have $\boldsymbol{C}^+$ scored higher than $\boldsymbol{C}^-$. 
We apply the pairwise cross-entropy loss to maximize the margin of $y^+$ and $y^-$ for $\boldsymbol{C}^+$ and $\boldsymbol{C}^-$, which equals to minimize:

\begin{equation}
\begin{aligned}
&\mathcal{L}_{Pred} = \log(1+\exp(y^{-}-y^{+})).
\end{aligned}
\end{equation}

\vspace{0.5em}
\noindent\textbf{Overall learning Objective.}
The three components of our model can be jointly optimized by minimizing the objective function:

\begin{equation}
\begin{aligned}
&\mathcal{L} = \mathcal{L}_{Pred} - \sum_t(\mathcal{L}_{Factor}^t),
\end{aligned}
\end{equation}

\noindent where $\mathcal{L}_{Factor}^t$ is for argument turn level.

\section{Experimental Setup}\label{sec:setup}

\vspace{0.5em}
\noindent\textbf{Data Preprocessing.}
We randomly split the dataset with $80$\% for training and $20$\% for test. 
Then, $20$\% of the training data is randomly selected for validation. 
For preprocessing, we take the the following steps. First, non-English terms were filtered out. Then, quotations, digits, and links were replaced with generic tags `$\langle$quote$\rangle$', `$\langle$digit$\rangle$', and `$\langle$url$\rangle$', respectively. Next, we employed the natural language toolkit (NLTK) for tokenization\footnote{\url{https://www.nltk.org/}}. After that, all letters were converted 
to lowercase. Finally, words occurred less than $10$ times were filtered out from the data.

\vspace{0.5em}
\noindent \textbf{Parameter Setting.}
We use Gated Recurrent Unit (GRU) as the RNN cell. The hidden size of GRU is set to $512$ with the word dropout rate of $0.2$. The dimensions of word embeddings and memory embeddings are both set to $200$. 
$\lambda=0.01$ following the setting of~\citet{DBLP:journals/tacl/abs-1903-07319} for balancing the MI loss.
For all the other hyperparameters, we tune them on the validation set by grid search. 
Optimization is performed using Adam~\cite{DBLP:journals/corr/KingmaB14}. In the learning process, we alternatively update the parameters of the argument factor encoder and the rest of our model. We run our model for $80$ epochs with early-stop strategy applied~\cite{DBLP:conf/nips/CaruanaLG00}.

\vspace{0.5em}
\noindent \textbf{Comparison Baselines.}
~\citet{DBLP:conf/www/TanNDL16} uses logistic regression with bag-of-words features in the pairwise pervasiveness prediction tasks, achieving good performance when compared with most of the handcrafted features.
Here we implement 
logistic regression with TfIdf-weighted $n$-grams features (\underline{\textsc{LR-Tfidf}}). 
Similar to~\cite{DBLP:conf/www/TanNDL16}, we adopt $\ell_1$ regularization on the training stage to avoid overfitting.
Joint topic-discourse model (\underline{\textsc{JTDM}})~\cite{DBLP:journals/tacl/abs-1903-07319} extracts topics and discourse features in an unsupervised way and can be used to place our argument factor encoder. 
We use the mean of each argument's topic-discourse mixture as the feature of an input conversation without considering the dynamics. 
Hierarchical attention recursive neural network (\underline{\textsc{HAtt-RNN}})~\cite{DBLP:conf/naacl/YangYDHSH16} uses bi-directional GRU as sequence encoder, including two levels of attention mechanisms (i.e., word level and argument level) while constructing the representation of a conversation.
Dynamic memory network (\underline{\textsc{DMN}})~\cite{DBLP:conf/icml/KumarIOIBGZPS16} is a neural sequence model that can encode the contextual history into the episodic memory component.
Dynamic key-value memory network (\underline{\textsc{DKVMN}})~\cite{DBLP:conf/www/ZhangSKY17} improves upon DMN using one static matrix as key to compute the memory reading weights and one dynamic matrix as value for updating the memory states. 

\section{Experimental Results}\label{sec:result}

This section presents the how models perform on persuasiveness prediction. We reports the main comparison results on persuasiveness prediction in Section \ref{ssec:pred_rst}, followed by topic and discourse interpretations in Section \ref{ssec:interprete_topic_disc}. 
Afterwards, we analyze the major parameters and errors in Section \ref{ssec:parameter-analysis} and Section~\ref{ssec:error-analysis} respectively.  

\subsection{Persuasiveness Prediction Comparison}\label{ssec:pred_rst}
We follow \citet{DBLP:conf/www/TanNDL16} to conduct pairwise classification. 
For the CMV dataset, we predict which conversation can win $\Delta$, and for the Court dataset, which side will win the case.
In Table~\ref{tab:pred_rst}, we report the pairwise accuracy and F1 scores.  For our models, we also display ablation results without considering topic, discourse, and memory structure, respectively. 
It is observed that:
\begin{table}[t]
	\center
	\caption{Pairwise classification results on persuasiveness prediction. 
Best results in \textbf{bold}.
Paired t-test is conducted between our full model and baselines/ablations ($^{\ast\ast}$: $p<0.01$,  $\ast$: $p<0.05$).} 
\vspace{-0.25cm}
	\scalebox{0.9}{
	\begin{tabular}{|l|ll|ll|}
		\hline
		\multirow{2}{*}{\textbf{Models}} & \multicolumn{2}{c|}{\textbf{CMV}} & \multicolumn{2}{c|}{\textbf{Court}} \\
		\cline{2-5}
		& Acc. & F1 & Acc. & F1 \\
		\hline
		\hline
		\underline{\textbf{Baselines}} &&&&\\
		\textsc{LR-Tfidf} & 0.571$^{\ast\ast}$ & 0.557$^{\ast\ast}$ & 0.631$^{\ast\ast}$ & 0.608$^{\ast\ast}$ \\
		\textsc{JTDM}  & 0.615$^{\ast\ast}$ & 0.586$^{\ast\ast}$ & 0.642$^{\ast\ast}$ & 0.625$^{\ast\ast}$ \\
		\textsc{HAtt-RNN} & 0.632$^{\ast\ast}$ & 0.641$^{\ast\ast}$ & 0.571$^{\ast\ast}$ & 0.538$^{\ast\ast}$ \\
	    \textsc{DMN} & 0.688$^{\ast\ast}$ & 0.673$^{\ast\ast}$ & 0.637$^{\ast\ast}$ & 0.602$^{\ast\ast}$ \\
	    \textsc{DKVMN} & 0.696$^{\ast\ast}$ & 0.698$^{\ast\ast}$ & 0.648$^{\ast\ast}$ & 0.629$^{\ast\ast}$ \\
		\hline
		\hline
		\underline{\textbf{DTDMN}} &&&&\\
		\textsc{w/o topic} & 0.713$^{\ast\ast}$ & 0.710$^{\ast\ast}$ & 0.653$^{\ast\ast}$ & 0.655$^{\ast\ast}$ \\
		\textsc{w/o discourse} & 0.749$^\ast$ & 0.745$^\ast$ & 0.671$^{\ast\ast}$ & 0.682$^{\ast\ast}$ \\
	    \textsc{w/o memory} & 0.707$^{\ast\ast}$ & 0.696$^{\ast\ast}$ & 0.616$^{\ast\ast}$ & 0.589$^{\ast\ast}$ \\
		\textsc{full model} & \textbf{0.751} & \textbf{0.748} & \textbf{0.702} & \textbf{0.694} \\
		\hline
	\end{tabular}
}

\vspace{-0.35cm}
\label{tab:pred_rst}
\end{table} 

\vspace{1mm}
\noindent$\bullet$~\textit{\textbf{Topic and discourse factors are useful.}}
By exploiting pre-learned latent topic and discourse factors, \textsc{JTDM} outperforms \textsc{LR-Tfidf} baseline on both datasets.
It even performs better than \textsc{HAtt-RNN} on Court debates. 
This observation implies that topic and discourse factors can be indicative of 
persuasiveness arguments.

\vspace{1mm}
\noindent$\bullet$~\textit{\textbf{Neural models generally outperform the non-neural baselines.}} 
This indicates that neural models are able to learn deep persuasiveness features. 
We also find that the improvement upon non-neural models is less significant 
on the Court compared to CMV. This may be partly attributed to the sparse training instances in the Court dataset 
as shown in Table \ref{tab:statistics}, 
which may result in overfitting. Nevertheless, our models can well alleviate such sparsity and achieve significantly better performance on both datasets.

\vspace{1mm}
\noindent$\bullet$~\textit{\textbf{Process modeling is important to predict argument persuasiveness.}} 
We observe that \textsc{LR-Tfidf} and \textsc{JTDM}, with only word features encoded, perform worse compared to 
other methods that explore dynamic patterns in argumentation process. 
This shows that persuasion outcomes are also dependent 
on a dynamic process beyond word features. 


\vspace{1mm}
\noindent$\bullet$~\textit{\textbf{Dynamic memory mechanism is effective.}} 
Our \textsc{full model} obtains better results than its \textsc{w/o memory} variant. 
Also, \textsc{DMN} and \textsc{DKVMN} outperform other baselines without dynamic memory mechanism. 
The above observations 
indicate that dynamic memory mechanism 
is effective for the argumentation progress.

\vspace{1mm}
\noindent$\bullet$~\textit{\textbf{Both dynamic topic and discourse factors contribute to argument persuasiveness.}} 
It is observed that our \textsc{full model} achieves better results than the \textsc{w/o topic} and \textsc{w/o discourse} ablation, which considers only dynamic discourse or topic factors.
Though the slightly better performance of \textsc{w/o discourse} than \textsc{w/o topic} shows that topic factors might contribute 
more to persuasiveness, coupling the topics and discourse exhibiting the best performance. 

\begin{table}[t]
\center
\caption{Top 10 representative words of example topics learned from CMV and Court. We manually label the topical content according to their associated words. }
\vspace{-0.3cm}
\scalebox{0.9}{
\begin{tabular}{|l|m{5.2cm}|}
\hline
\textbf{Topic} & \textbf{Words} \\  
\hline
\hline
\textit{Academic (CMV)} & phd sociology genetics predetermined biology quantum field classical influences refers \\ 
\hline
\textit{Foreign culture (CMV)}  & japanese europeans european spanish french africans german indian native heritage \\
\hline
\textit{Ecocrisis (CMV)} & chernobyl fukushima warming nirvana tolerance hydroelectric dangers grunge swastika warnings \\
\hline
\textit{Criminal histroy (Court)} & juvenile evasion adult youth records olds history sims thorough court \\
\hline
\textit{Commerce (Court)} & income billion lawful cents marketplace revenue supply descriptive rate transactions \\
\hline
\textit{Political party (Court)} & voting voters attract republican republicans challenger democrats poaching vote candidate \\
\hline
\end{tabular}
}

\vspace{-0.3cm}
\label{tab:topic_sample}
\end{table}


\subsection{Interpretation on Topics and Discourse}\label{ssec:interprete_topic_disc}

To analyze the latent topics and discourse produced our model, we carry out a qualitative analysis to investigate their interpretability.
Here we select the top $10$ words from the distributions of some example topics and discourse factors and list them in Table~\ref{tab:topic_sample} and Table~\ref{tab:disc_sample}, respectively. 
It can be observed that there are some meaningful word clusters reflecting varying debate topics and discourse skills on the two datasets. Interestingly, we observe that latent discourse from CMV and Court, though learned separately, exhibit some overlaps in their corresponding top 10 words; particularly for ``pronoun'', which are used to refer to participants (e.g., ``we'') or someone/something else (e.g., ``he'' and ``they'') in the discourse.
We also note that the discourse style of ``statistic'' is represented by very different words. The reason is that Court debates are likely to involve lawsuit-related statistical evidence, hence exhibiting the prominence of words like ``records'' and ``proximate''.

\begin{table}[t]
\center
\caption{Top 10 representative words of example discourse styles learned from CMV and Court. The discourse styles of the word clusters are manually assigned according to their associated words. }
\vspace{-0.25cm}
\scalebox{0.9}{
\begin{tabular}{|l|m{3cm}|m{3cm}|}
\hline
\textbf{Discourse Role} & \textbf{CMV} & \textbf{Court} \\  
\hline
\hline
\textit{Question}  & what <quote> ? why were how would has could our & can ? you if further him are we go take \\ 
\hline
\textit{Pronoun}   & my he him his we one am on 've myself & his he was no after did they not this became \\
\hline
\textit{Conjecture}& as there per more from less often ). than many  & no there i further don either your any many be \\
\hline
\textit{Quotation} & [ ' <url> \string^[[ )][/ )][[ \string^| (~< >:< :]( & .] '' i don if you find would .) \\
\hline
\textit{Statistic} & more <digit> was than less from could been ' had & resulting <digit> regs records sims definitional proximate thorough counts instruction \\
\hline
\end{tabular}
}

\vspace{-0.25cm}
\label{tab:disc_sample}
\end{table}
\begin{figure*}[t]
	\centering
	\includegraphics[width=0.99 \textwidth]{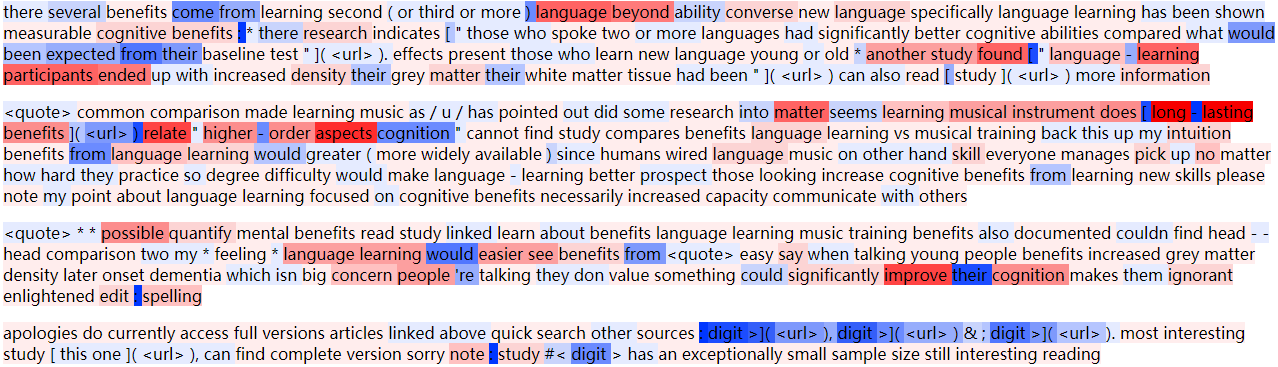}
	\caption{
    Visualization of the topic-discourse assignment of CMV conversation in Figure \ref{fig:intro-example}. The annotated blue words are pone to be discourse words, and the red are topic words. The shade is the word-level confidence of current assignment.
    }
	\label{fig:case-td}
\end{figure*}

We further explore why coherent topics and discourse styles can be learned with the example conversation in Figure~\ref{fig:intro-example}.
In Figure~\ref{fig:case-td}, we visualize the topics and discourse assignments where we highlight the topic words (with $p(w|\boldsymbol{z}) > p(w|\boldsymbol{d})$) in red, the rest in blue to indicate discourse style. 
The shade indicates the confidence level of such word assignment. 
We can see that our model can identify the topic words, e.g., ``\textit{language beyond}'', ``\textit{found}'', and ``\textit{learning participants ended}'', and also discourse words, e.g., ``$<$digit$>$'' and ``[''.
It is seen that topics and discourse words can be well distinguished, which allows us to discover meaningful latent factors and analyze reasons behind persuasive arguments.

\subsection{Parameter Analysis} \label{ssec:parameter-analysis}
Here we study how the two important hyper-parameters in our model, the number of topics ($K$) and discourse ($D$) affect our model performance. In Figure \ref{fig:topic_disc_num}, we show the persuasiveness prediction accuracy given varying $K$ in (a) and varying $D$ in (b). 

As can be seen, for both topic and discourse, the curves corresponding model performance are not monotonic. 
In particular, better accuracies are achieved given relatively larger topic numbers for CMV with the best result observed at $K=50$. 
While for Court, the optimum topic number is $K=20$.
This may be due to 
the relatively more centralized topics in Court debates, whereas wider range of topics discussed in social media, CMV.
For discourse, we observed a similar trend in both CMV and Court datasets. 
The best score is achieved when $D=10$ for CMV and $D=8$ for Court dataset.
This implies that 
the discourse styles used in both CMV and Court are somewhat limited. 


\subsection{Error Analysis} \label{ssec:error-analysis}
In this section, we look into the errors produced by our model 
in predicting the argumentation persuasiveness, where three types of major errors are observed. 

\begin{figure}[t]
	\centering
	\begin{subfigure}[h]{0.23\textwidth}
	\includegraphics[width=0.98 \textwidth]{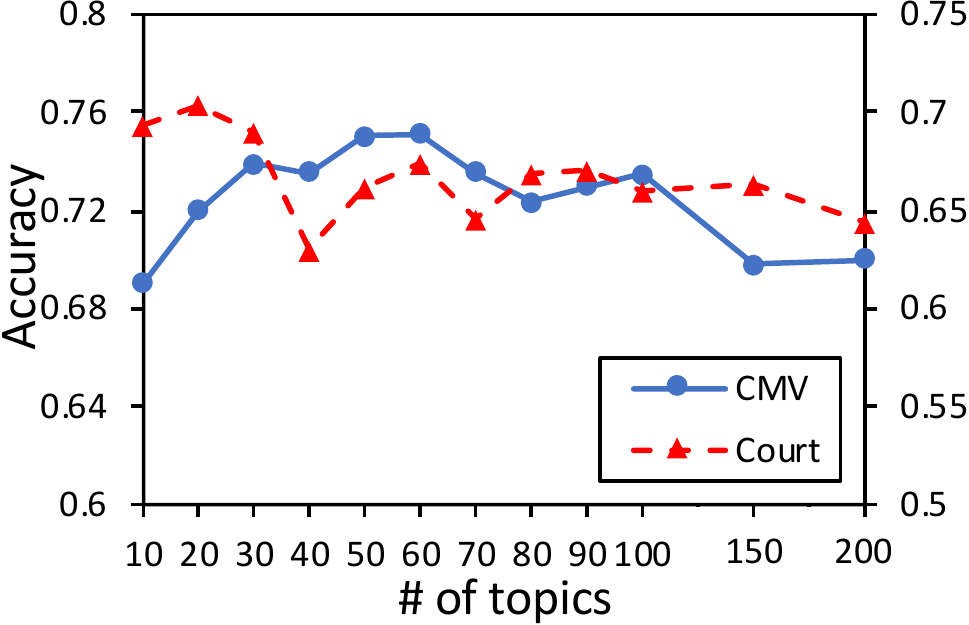}
    \caption{Effect of topic number}
    \end{subfigure}
    \hfill
    \begin{subfigure}[h]{0.23\textwidth}
    \includegraphics[width=0.98 \textwidth]{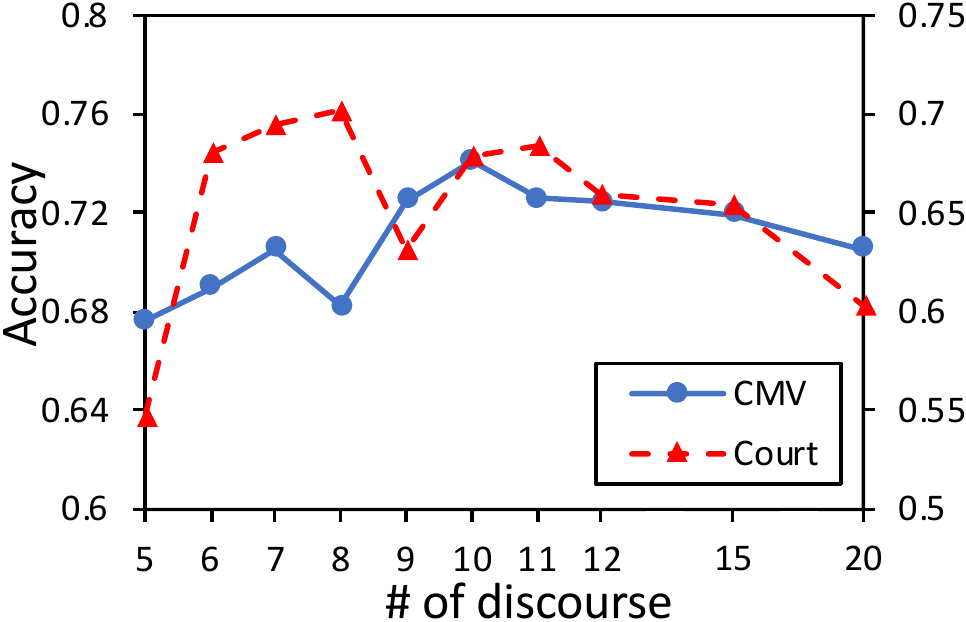}
    \caption{Effect of discourse number}
    \end{subfigure}
    
    \centering
    \caption{The impact of topic number (a) and discourse number (b) on our model for persuasiveness prediction. 
    For both (a) and (b), the blue and solid line shows the results on CMV with left vertical axis, and the red and dashed line Court with the right vertical axis.
    }
    \vspace{-0.25cm}
\label{fig:topic_disc_num}
\end{figure}

\vspace{0.2em}
\noindent\textbf{Error Type I: Wrong separation of the topic words and discourse words.} The errors occur in distinguishing topic and discourse words may result in erroneous 
persuasiveness prediction results. 
For example, as shown in Figure~\ref{fig:topic_disc_num}, the word ``\textit{cognition}'' should be considered as a topic word yet erroneously inferred to reflect discourse
Because of the cascading failure, the model output might be affected.

\vspace{0.2em}
\noindent\textbf{Error Type II: 
Preconception held by opinion holders.} 
Sometimes opinion holders hold preconception towards the debating subject and their views are difficult to be changed by others. 
As shown in Figure~\ref{fig:error_case_2}, the opinion holder raised an issue related to the ``\textit{abortion ban act}''. Although the challengers provide arguments with concrete evidences against the $OP$, they fail to obtain a $\Delta$ due to the opinion holder's preconception. 
Such cases are prominent on social media, posing a challenge to understand opinion holders' prior beliefs for a better prediction of argumentation outcome.

\vspace{0.2em}
\noindent\textbf{Error Type III: Lack of knowledge for judging 
the sufficiency of the evidence.} In the court scenario, successful debates depend on how the lawyers make use of their persuasive skills to present their evidence or interpret their opponents' evidence. The judgement of the evidence sufficiency is beyond what the current model can capture. 
The logic and sufficiency of evidence could not be easily determined without external knowledge, e.g., law terminology and clauses. In future work, we will strengthen the reasoning process of the model by incorporating external knowledge sources.

\section{Discussion}\label{sec:discussion}

In Section \ref{sec:result}, we have shown 
the superior performance of our proposed model to identify persuasiveness arguments. 
Here, we discuss how the latent topics and discourse signal argumentation outcome. 
From our results, we further draw three suggestions, which might help individuals better engage in argumentative dialogues.

\subsection{The Roles of Topics and Discourse in Argumentation Process}

In Section~\ref{ssec:pred_rst}, topics have shown slightly stronger effects on successful persuasion than discourse. 
Here we further analyze their 
roles in affecting persuasion outcome. 
Similar trends are observed on both datasets and we only discuss the results on CMV dataset due to the space limitation.

\begin{figure}[t]
    \centering
    \small
    \begin{tabular}{|m{7cm}|}
    \hline
    \textbf{Title:} Abortion is not a woman's rights issue ...\\
    \textbf{OP:} I have never met a person who said they were against abortion because they did not think woman should have autonomy over their own body. ...\\
    \textbf{Wining Side:} ... When one right is a person's right to be alive, I think the woman's right is unimportant. ...\\
    \textbf{Losing Side:} ... Abortion is fundamentally a ``prioritizing'' of rights. The rights of the fetus vs. the rights of the woman. ... <quote> No. We can't know what the world would be like if men could get pregnant. ...\\
    
    
    \hline
    \end{tabular}
    \caption{An example of inaccurate prediction caused by the preconception of the opinion holder. The losing side put forward stronger argument and far more evidences compared to the winning side (15 vs. 5 sentences). 
    We only show partial arguments here due to the space constraint. }
    \label{fig:error_case_2}
    \vspace{-0.25cm}
\end{figure} 

\begin{figure}[t]
	\centering
	\begin{subfigure}[h]{0.23\textwidth}
    \centering
	\includegraphics[width=0.98 \textwidth]{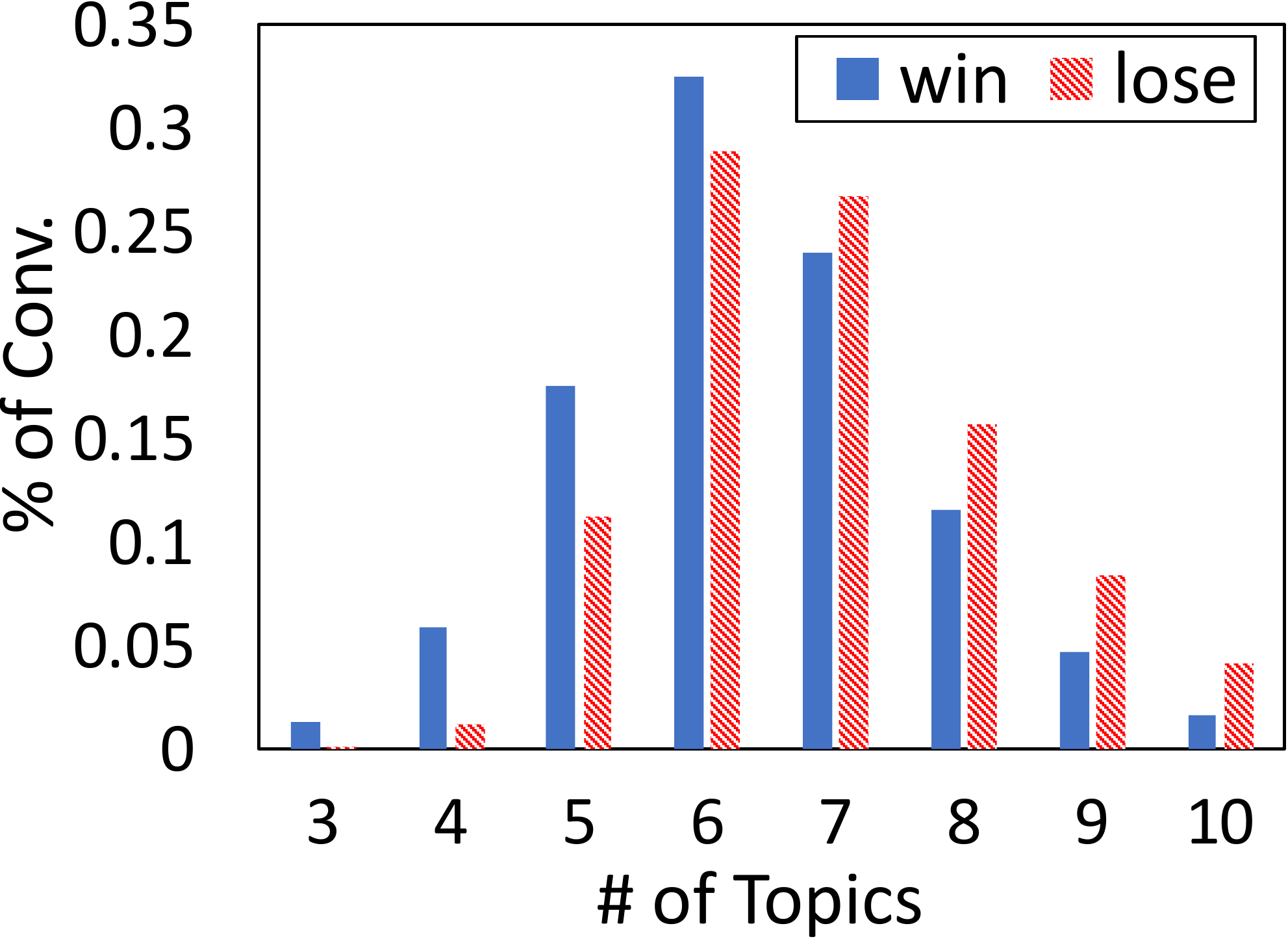}

	\caption{Topic}
    \end{subfigure}
    \begin{subfigure}[h]{0.24\textwidth}
    \centering
    \includegraphics[width=0.98 \textwidth]{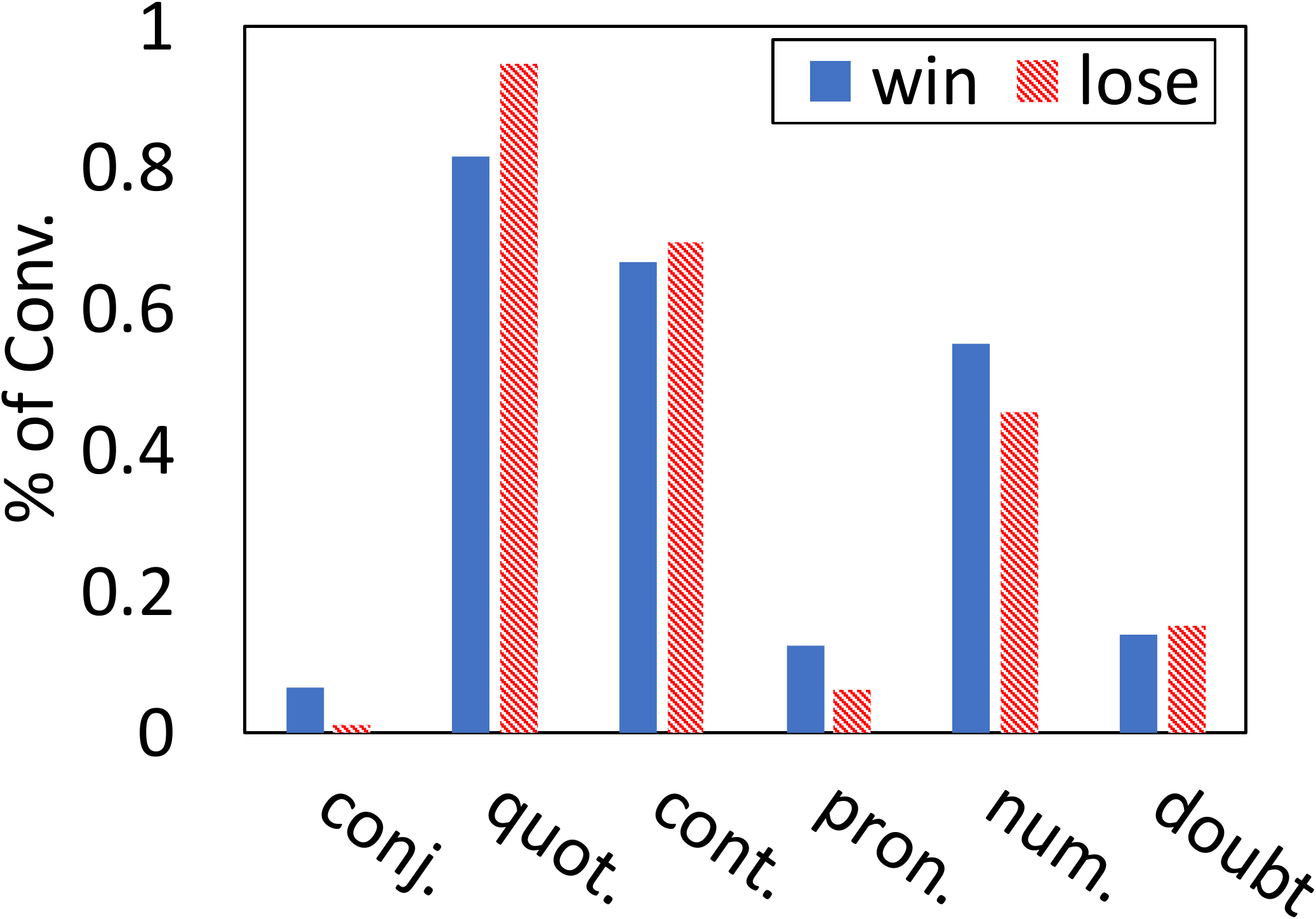}

	\caption{Discourse}
    \end{subfigure}

	\caption{
	Distributions of winning and losing persuasion over the number of \emph{strong} argument topics involved in (a) and varying discourse factors in (b).
	For (b), we display the discourse factors with our interpretation (``conj.''-conjunction, ``quot.''-quotation, ``cont.''-contrast, ``pron.''-personal pronoun, and ``num.''-statistic).
    Two-sided Mann-Whitney rank test shows that the two distributions 
    are significantly different for both sides ($p<0.01$).
    }
    \vspace{-0.5cm}
	\label{fig:topic_dis}
\end{figure}

To investigate topic effects, we follow \citet{DBLP:journals/tacl/WangBSQ17} to identify \emph{strong} argument topics when the topic likelihood is larger than a pre-defined threshold (set to $0.2$ here).\footnote{To set the threshold, we first sample $100$ arguments and manually align them to the strongly related latent topic. 
Then we set the threshold resulting in the smallest errors compared with human annotations.} Then in Figure \ref{fig:topic_dis}(a), we show how the number of strong argument topics distribute over winning arguments compared with the losing ones. 
For discourse, we similarly show the discourse factor distributions on winning and losing arguments in Figure \ref{fig:topic_dis}(b). Here we display the discourse factors with our interpretations on the discourse styles according to their associated word distributions.
In the following we discuss the findings from topics and discourse in turn.

\vspace{0.2em}
\noindent\textbf{Topic Roles.} As can be seen in Figure~\ref{fig:topic_dis}(a), the winning side tends to put forward fewer topics in the argumentative process. 
This indicates that 
strong and focused argument points are more closely related to successful arguments than diverse topics, because arguing with
too many things 
might overwhelm the opinion holder, which may lead to the persuasion failure.
Similar trend can also be observed in Court dataset.

\vspace{0.2em}
\noindent\textbf{Discourse Roles.} From Figure~\ref{fig:topic_dis}(b), we can see discourse styles vary in their 
effects over the persuasiveness results. 
Specifically, personal pronoun and statistic are more likely to appear in the winning side than the losing side.
Their positive effects have also been previously reported~\cite{DBLP:conf/www/TanNDL16,DBLP:journals/tacl/WangBSQ17}.
Moreover, we find that conjunction, though not widely used, is obviously more endorsed by winning sides.
The benefit of conjunction may 
be related with the better logic it renders.
For the losing side, they are more in favor of the quotation discourse, which is used in CMV to quote and attack others' weak points.
People may dislike such criticism, which renders the negative impact on persuasiveness.

\subsection{Discourse Effects over Turn Number}

\begin{figure}[t]
	\centering
	\begin{subfigure}[h]{0.23\textwidth}
	\includegraphics[width=0.98 \textwidth]{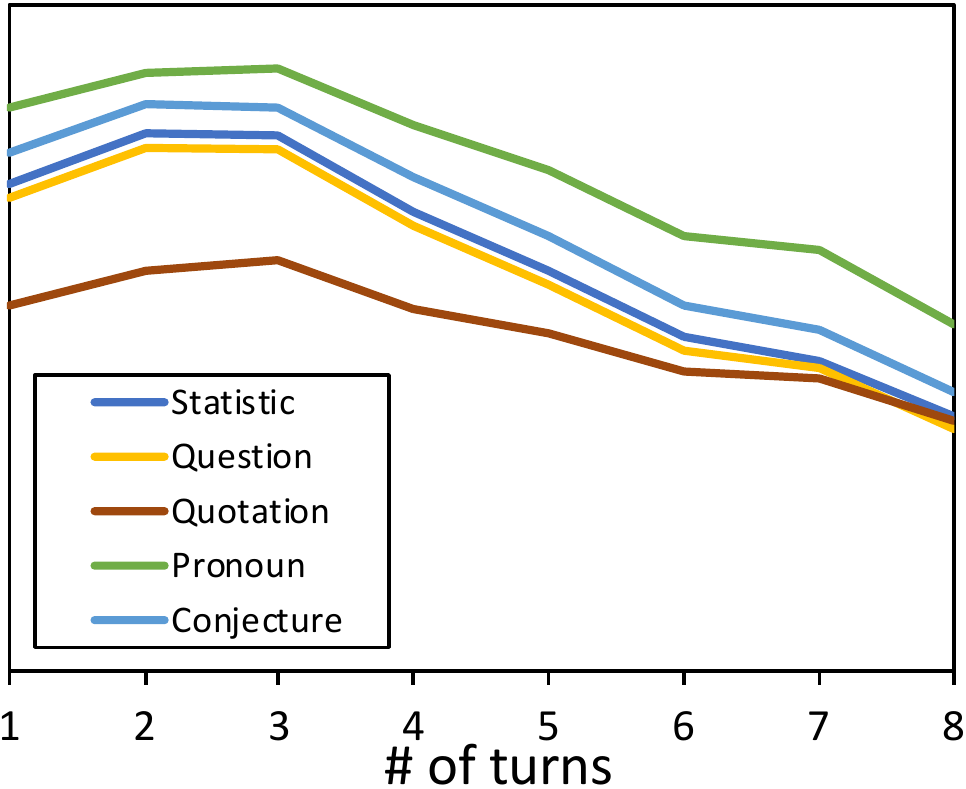}
    \caption{CMV}
    \end{subfigure}
    \hfill
    \begin{subfigure}[h]{0.23\textwidth}
    \includegraphics[width=0.98 \textwidth]{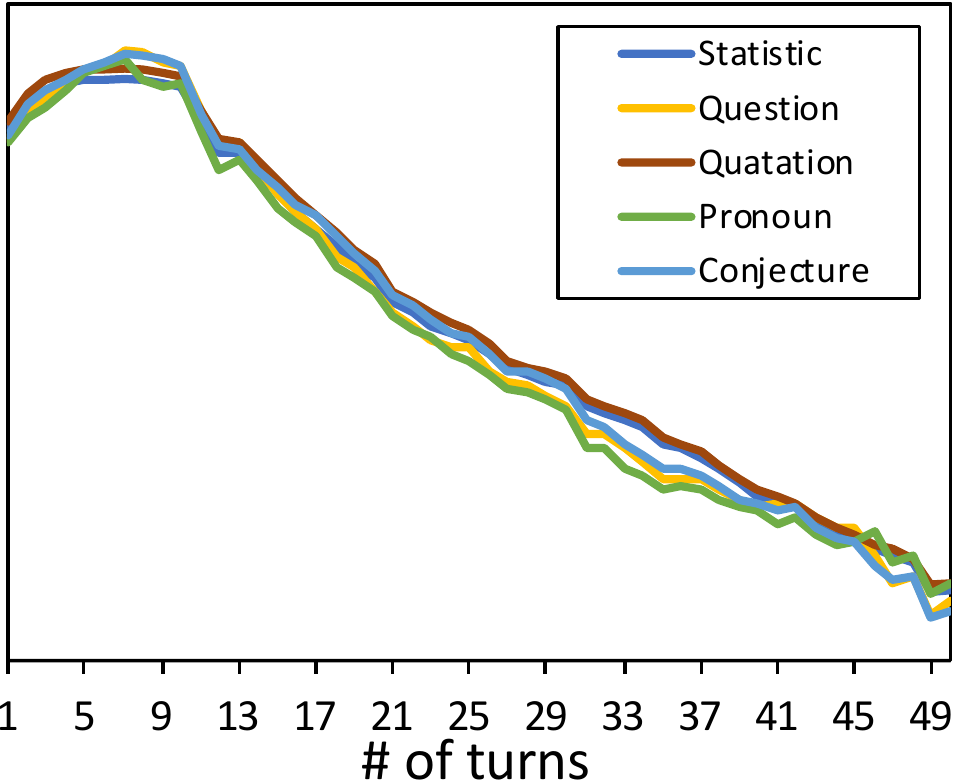}
    \caption{Court}
    \end{subfigure}
    
    \centering
    \caption{Effects of the example discourse styles (in Table \ref{tab:disc_sample}) on the ultimate persuasiveness as the argumentation process continues. The horizontal axis indicates the number of argumentative turns, and the vertical axis the dynamic persuasiveness score (given by Eq.~\ref{eq:pred_score}). 
    }
    \vspace{-0.25cm}
\label{fig:disc_turns}
\end{figure}

To provide more insights, we further study the change of discourse effects over argumentation process with varying conversation length (the number of turns).
Here we focus on discourse effects instead of topics because discourse styles are commonly used in diverse arguments and exhibit shared patterns on the two dataset, while topics vary in different scenarios. 
Specifically, we investigate the effects of the example discourse (listed in Table~\ref{tab:disc_sample}) over argumentation processes with varying turn number. 
The persuasiveness scores (computed with Eq.~\ref{eq:pred_score}) are employed to measure the discourse effects and the results on the two datasets are displayed in Figure~\ref{fig:disc_turns}. Here comes our observations.

First, we find that in general all the discourse styles exhibit 
a decreasing trend in terms of persuasiveness scores with more argumentative turns coming in, 
although they appear to be more important in the initial few turns. 
Second, same discourse style may demonstrate varying persuasiveness impacts in different debate scenarios. For example, the pronouns usually express more personal emotion and tend to arouse empathy from others. Such discourse shows a positive effect on debates in the social media scenario, as shown in Figure~\ref{fig:disc_turns} (a), but its effectiveness in the Court scenario is less apparent. A similar observation can be found 
for the quotations. Finally, we also observe that on the Court dataset, various 
discourse styles exhibit very similar effects on the persuasion, as shown in Figure~\ref{fig:disc_turns} (b), while in CMV, the effects of various discourse styles can be easily distinguished.

\subsection{Case Study
}\label{ssec:topic_disc_effect}

\begin{figure*}[h]
	\centering
	\includegraphics[width=0.95 \textwidth]{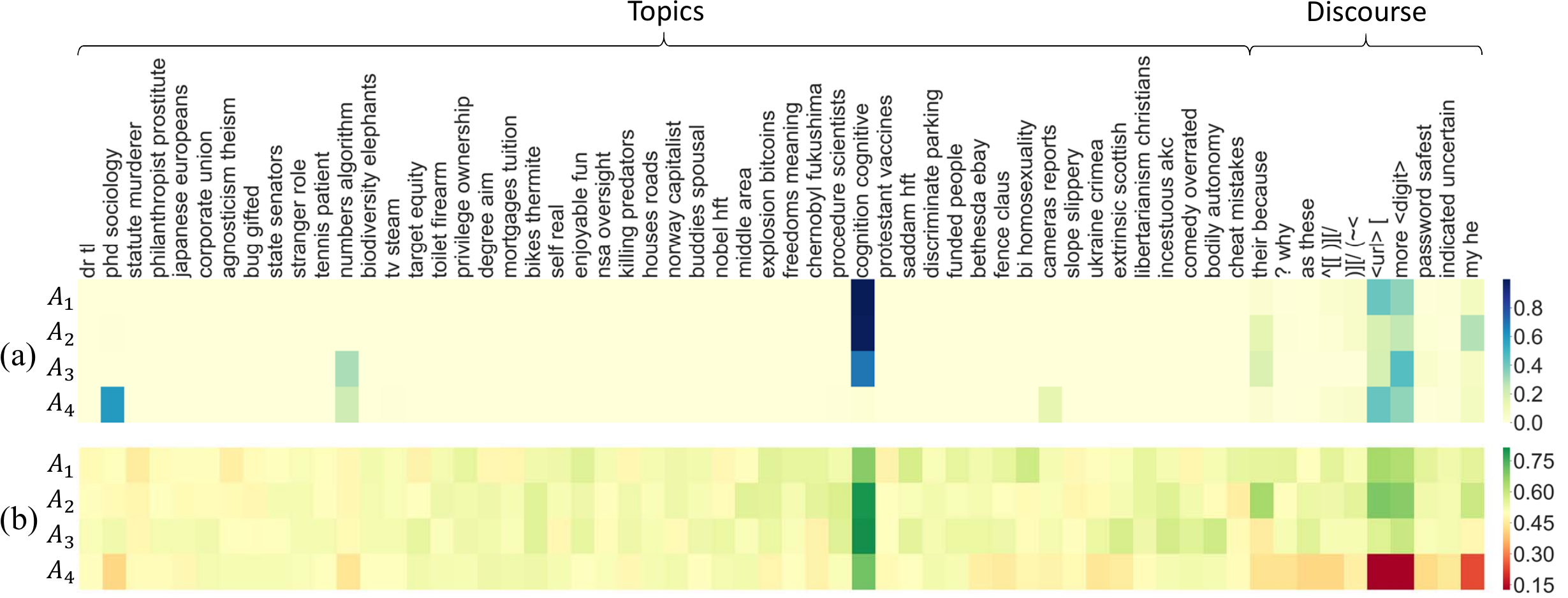}
	\vskip -0.5em
	\caption{The heatmap visualizing the dynamic memory weights and persuasiveness on topic and discourse factors for the conversation in Figure \ref{fig:intro-example}. 
	The vertical axis shows the turn id (from $A_1$ to $A_4$), and horizontal axis shows the latent topics and discourse displayed with their top $2$ words.
	(a) Dynamic memory weights $\boldsymbol{w}^t$ that indicate topics shift and discourse flow. 
	(b) Persuasiveness effect from each topic or discourse.
	For (a) and (b), darker colors indicate higher impacts. 
	For (b), green indicate positive impacts while red negative.
	}
\label{fig:topic_disc}
\vskip -0.5em
\end{figure*}

Our DTDMN is designed for capturing the topic shifting and discourse flow in an argumentation conversation, which allows us to interpret argument persuasiveness from the perspectives of topic and discourse dynamics. 
Here we take the CMV discussion in Figure~\ref{fig:intro-example} as an example to look into its persuasion process.
Recall that the challengers put forward viewpoints centered around ``\emph{the advantage to learn a second language}", and they successfully change the opinion holder's mind with good arguments delivered.
In Figure~\ref{fig:topic_disc}(a), we visualize the dynamic memory weights $\boldsymbol{w}^t$ (see Eq.~\ref{eq:weight}) for each turn.
It is observed that our model highlights the  `\textit{cognition}' topic factor, which suggests the cognitive research evidence (e.g., learning a musical instrument) might help challengers win.
For discourse, the model 
highlights latent factors represented by words like `$\langle$url$\rangle$', `$\langle$digits$\rangle$', and `\textit{more}'.
This suggests that effective discourse styles, such as 
quotation of links (`$\langle$url$\rangle$') and statistic (`$\langle$digits$\rangle$'), may also play an important role in persuasiveness.

To further study how each topic and discourse alone contributes to this example's persuasion, we disable the effects from other topics and discourse via masking $\boldsymbol{w}^t$, and map the prediction score $y$ in Eq.~\ref{eq:pred_score} to $[0,1]$ range. We visualize the prediction scores in Figure~\ref{fig:topic_disc}(b) to depict the effect of persuasiveness from each topic and discourse.
We observe that the ``\textit{cognition}'' topic is still highlighted for all turns. 
It implies our model still recognizes this topic to be important, without taking the discourse effects into account. 
For discourse, we notice that the \textit{quotation} and \textit{statistic} skills are considered useful for the first few turns, whose impacts however later change to be negative. 
It might be because people tend to be tired of excessive URL links and statistics
without providing more insightful opinions. 

\subsection{Suggestions on Argumentation}
From the results, we draw some general suggestions on argumentation, which might help participants behave better in debates.

\vspace{0.2em}
\noindent\textbf{Topics are more important than discourse styles.} In an argumentative conversation, opponents attempt to establish the validity of two positions by convincing each other and trying to win points in the debate~\cite{stein2001origins}. Our study shows that topics contribute slightly more on persuasiveness than discourse.
It happens especially in later stage of the argumentation process, which is suggested by the decreasing effects of discourse over turn number (see Figure \ref{fig:disc_turns}).
This is consistent with the discovery in ~\citet{van1983strategies}, which points out that style
and rhetoric are not the dominant factors to determine debate outcome.

\vspace{0.2em}
\noindent\textbf{Strong and focused argument points are better than diverse topics.} Strong arguments that are well-supported with evidence and/or reasoning, generally deliver more persuasive messages to audience~\cite{benoit1987argumentation}. Our study reveals that successful argumentation usually conveys fewer and focused topics. Diverse topics could only distract audience and expose more vulnerable points to the opponent.

\vspace{0.2em}
\noindent\textbf{Well organize the points and address them in a modest and concrete way}. 
Argument discourse represents the cultural and situational realities of human reasoning, and is more sensitive to audience in conversational debates~\cite{ellis2015argument}. \citet{amossy2009argumentation} also claims that argumentativity constitutes an inherent feature of discourse.
This advice works particularly well in social media arguments, where the amateur debaters from general public are likely to be affected by opponents' discourse skills.
As a result, we see that personal pronoun (modest), statistic (concrete), and conjunction (well-organized) discourse are more likely to appear in wining side.


\subsection{Limitations of Our Study}\label{ssec:threat}

In this paper, we use the CMV dataset following the previous augmentation mining setting~\cite{DBLP:conf/www/TanNDL16,DBLP:conf/coling/JiWHLZH18,DBLP:conf/aaai/HideyM18,DBLP:conf/argmining/LinHHC19}. 
In them, only the CMV dataset is used in evaluation. 
To better evaluate the 
generalization performance, 
we also include the Supreme Court dataset, which exhibits different data statistics from CMV (e.g., fewer argumentation processes and more turns involved in a process). 
However, it might not guarantee the generalization capability of our proposed model on other argumentation genres. 
Also, our findings are drawn 
from the experimental results of the CMV and the Court datasets.
These findings are consistent with prior studies in social linguistics, and provide some additional details. 
To further evaluate if our model and empirical results are applicable in other scenarios, more experimental study is required on a diverse range of debate data to better understand human arguments.

In addition, we mainly consider the topic and discourse factors 
in the modeling of the argumentation process. There are other factors that may relate to the persuasiveness of an argumentative conversation, such as age~\cite{felton2004development}, culture~\cite{tirkkonen1996explicitness},  gender~\cite{Jeong2006} of participants. 
For example, earlier research~\cite{felton2001development} on argumentation suggests that adults use advanced discourse strategies more consistently, frequently, and flexibly than adolescents do. Due to the unavailability of such metadata in 
our datasets, we could not easily incorporate these factors. Future research can consider building debate datasets with side information such as demographics data included.

\section{Conclusion and Future Work}\label{sec:con}
In this paper, we propose to dynamically track both topics and discourse factors in conversational argumentation for persuasiveness prediction. The proposed neural model not only identifies persuasive arguments more accurately, but also provides insights into the usefulness of topics and discourse for a successful persuasion. The findings concluded in this paper can facilitate the argument persuasiveness analysis.


\section*{Acknowledgements}
The work described in this paper is supported by the Research Grants Council of the Hong Kong Special Administrative Region, China (No. CUHK 14210717 of the General Research Fund and No. CUHK 2410021 of the Research Impact Fund R5034-18). Jing Li is supported by the Hong Kong Polytechnic University Internal Fund (1-BE2W). YH is partly funded by the EPSRC grant EP/T017112/1. 


\bibliographystyle{ACM-Reference-Format}
\bibliography{main}


\end{document}